\RequirePackage{fix-cm}
\documentclass[smallextended]{svjour3}       % onecolumn (second format)
\smartqed  % flush right qed marks, e.g. at end of proof
\usepackage{graphicx}
\usepackage{amsmath,amssymb,url,cite}
\def\bo{\mbox{\boldmath$o$}}

\begin{document}

\title{Neuro-SERKET: Development of Integrative Cognitive System through the Composition of Deep Probabilistic Generative Models}

\titlerunning{Neuro-SERKET: Development of Integrative Cognitive System through the Composition of PGMs}        % if too long for running head

\author{Tadahiro Taniguchi \and \\
        Tomoaki Nakamura \and
        Masahiro Suzuki \and \\
        Ryo Kuniyasu \and
        Kaede Hayashi \and \\
        Akira Taniguchi \and
        Takato Horii \and \\
        Takayuki Nagai
}


\institute{T. Taniguchi, A. Taniguchi, and K. Hayashi \at
              Ritsumeikan University, 1-1-1 Noji-higashi, Kusatsu, Shiga \\
              Tel.: +81-77-561-5745\\
              Fax:  +81-77-561-5745\\
              \email{\{taniguchi, a.taniguchi, k.hayashi\}@em.ci.ritsumei.ac.jp}           %  \\
           \and
           T. Nagai, and T. Horii \at
             Osaka University, 1-3 Machikane-yama, Toyonaka, Osaka \\
              Tel.: +81-6-6850-6365\\
              Fax:  +81-6-6850-6365\\
              \email{\{nagai, takato\}@sys.es.osaka-u.ac.jp}           %  \\
           \and
           T. Nagai, T. Nakamura, and R. Kuniyasu \at
           The University of Electro-Communications, 1-5-1 Chofugaoka, Chofu, Tokyo \\
           Tel.: +81-42-443-5215\\
           Fax.: +81-42-443-5215\\
           \email{\{tnakamura@, r\_kuniyasu@radish.ee.\} uec.ac.jp }
           \and
            M. Suzuki \at
            The University of Tokyo,  7-3-1, Hongo,Bunkyo-ku, Tokyo\\
            Tel.: +81-3-3815-5411\\
            Fax:  +81-3-3815-5411\\
            \email{masa@weblab.t.u-tokyo.ac.jp}
}

\date{Received: date / Accepted: date}

\maketitle

\begin{abstract}
This paper describes a framework for the development of an integrative cognitive system based on probabilistic generative models (PGMs) called  Neuro-SERKET. Neuro-SERKET is an extension of SERKET, which can compose elemental PGMs developed in a distributed manner and provide a scheme that allows the composed PGMs to learn throughout the system in an unsupervised way. In addition to the head-to-tail connection supported by SERKET, Neuro-SERKET supports tail-to-tail and head-to-head connections, as well as neural network-based modules, i.e., deep generative models. As an example of a Neuro-SERKET application, an integrative model was developed by composing a variational autoencoder (VAE), a Gaussian mixture model (GMM), latent Dirichlet allocation (LDA), and automatic speech recognition (ASR). The model is called VAE+GMM+LDA+ASR. The performance of VAE+GMM+LDA+ASR and the validity of Neuro-SERKET were demonstrated through a multimodal categorization task using image data and a speech signal of numerical digits.      
\end{abstract}

\section{Introduction}\label{sec:1}
\def\bq{\begin{equation}}
\def\eq{\end{equation}}
\def\beq{\begin{eqnarray}}
\def\eeq{\end{eqnarray}}
\def\ba{\begin{array}}
\def\ea{\end{array}}
\def\bc{\begin{center}}
\def\ec{\end{center}}

\def\dsum{\sum\limits}
\def\disp{\displaystyle}
\def\ejw{e^{j\omega}}
\def\ejwi{e^{j\omega_{i}}}
\def\e-jwi{e^{-j\omega_{i}}}
\def\dfrac#1#2{\disp{\frac{#1}{#2}}}
\def\teigi{\stackrel{\triangle}{=}}

\def\b0{\bf{0}}
\def\bPhi{\mbox{\boldmath$\Phi$}}
\def\baa{\mbox{\boldmath$a$}}
\def\bb{\mbox{\boldmath$b$}}
\def\bcc{\mbox{\boldmath$c$}}
\def\bd{\mbox{\boldmath$d$}}
\def\be{\mbox{\boldmath$e$}}
\def\bff{\mbox{\boldmath$f$}}
\def\bg{\mbox{\boldmath$g$}}
\def\bh{\mbox{\boldmath$h$}}
\def\bi{\mbox{\boldmath$i$}}
\def\bj{\mbox{\boldmath$j$}}
\def\bk{\mbox{\boldmath$k$}}
\def\bl{\mbox{\boldmath$l$}}
\def\bm{\mbox{\boldmath$m$}}
\def\bn{\mbox{\boldmath$n$}}
\def\bo{\mbox{\boldmath$o$}}
\def\bp{\mbox{\boldmath$p$}}
\def\bqq{\mbox{\boldmath$q$}}
\def\br{\mbox{\boldmath$r$}}
\def\bs{\mbox{\boldmath$s$}}
\def\bt{\mbox{\boldmath$t$}}
\def\bu{\mbox{\boldmath$u$}}
\def\bv{\mbox{\boldmath$v$}}
\def\bw{\mbox{\boldmath$w$}}
\def\bx{\mbox{\boldmath$x$}}
\def\by{\mbox{\boldmath$y$}}
\def\bz{\mbox{\boldmath$z$}}

\def\bA{\mbox{\boldmath$A$}}
\def\bB{\mbox{\boldmath$B$}}
\def\bC{\mbox{\boldmath$C$}}
\def\bD{\mbox{\boldmath$D$}}
\def\bE{\mbox{\boldmath$E$}}
\def\bF{\mbox{\boldmath$F$}}
\def\bG{\mbox{\boldmath$G$}}
\def\bH{\mbox{\boldmath$H$}}
\def\bI{\mbox{\boldmath$I$}}
\def\bJ{\mbox{\boldmath$J$}}
\def\bK{\mbox{\boldmath$K$}}
\def\bL{\mbox{\boldmath$L$}}
\def\bM{\mbox{\boldmath$M$}}
\def\bN{\mbox{\boldmath$N$}}
\def\bO{\mbox{\boldmath$O$}}
\def\bP{\mbox{\boldmath$P$}}
\def\bQ{\mbox{\boldmath$Q$}}
\def\bR{\mbox{\boldmath$R$}}
\def\bS{\mbox{\boldmath$S$}}
\def\bT{\mbox{\boldmath$T$}}
\def\bU{\mbox{\boldmath$U$}}
\def\bV{\mbox{\boldmath$V$}}
\def\bW{\mbox{\boldmath$W$}}
\def\bX{\mbox{\boldmath$X$}}
\def\bY{\mbox{\boldmath$Y$}}
\def\bZ{\mbox{\boldmath$Z$}}
\def\bomega{\mbox{\boldmath$\omega$}}
\def\bLambda{\mbox{\boldmath$\Lambda$}}
\def\blambda{\mbox{\boldmath$\lambda$}}
\def\bmu{\mbox{\boldmath$\mu$}}
\def\bSigma{\mbox{\boldmath$\Sigma$}}
\def\bphi{\mbox{\boldmath$\phi$}}
\def\bPhi{\mbox{\boldmath$\Phi$}}
\def\balpha{\mbox{\boldmath$\alpha$}}
\def\bTheta{\mbox{\boldmath$\Theta$}}
\def\btheta{\mbox{\boldmath$\theta$}}
\def\bGamma{\mbox{\boldmath$\Gamma$}}
\def\bPsi{\mbox{\boldmath$\Psi$}}
\def\bDelta{\mbox{\boldmath$\Delta$}}
\def\bPi{\mbox{\boldmath$\Pi$}}

\makeatletter
\def\lddots{\mathinner{\mkern1mu\raise\p@\vbox{\kern7\p@\hbox{.}}\mkern2mu
\raise4\p@\hbox{.}\mkern2mu\raise7\p@\hbox{.}\mkern1mu}}
\makeatother

\def\argmax{\mathop{\rm argmax}}

The development of integrative cognitive systems that can form perceptual and behavioral concepts using multimodal sensorimotor information and learn and understand a language in a real-world environment is a significant challenge in artificial intelligence (AI) and robotics~\cite{Taniguchi2018symbol}. This paper describes a theoretical framework called Neuro-SERKET for this purpose.

Numerous types of integrative cognitive systems, which are sometimes called a cognitive architecture, have recently been developed for building service robots and modeling human adaptive cognition~\cite{nakamuraIROS07,nishide2008,ogata2010inter,mangin2015mca,sinapov2014grounding,miyazawa2018,taniguchi2017online,Taniguchi2016symbol,tani2016exploring,nakamura2017serket}. However, the cognitive systems for robots need to handle a variety of types of sensorimotor modalities, e.g., image, sound and actuation, and a variety of internal cognitive processes, e.g., categorization and planning. Therefore, the size of the computational models becomes large and the development requires significant effort for each integrative cognitive system. For further progress in this stream of research, we need to achieve an efficient way to develop complex cognitive systems in a practical manner. In addition, recent advancements in deep generative models (DGMs), for instance, a variational auto-encoder (VAE) ~\cite{kingma2013auto}, have boosted their utilization in the development of cognitive systems.

\begin{figure}[t]
 \centering
 \includegraphics[width=0.8\linewidth]{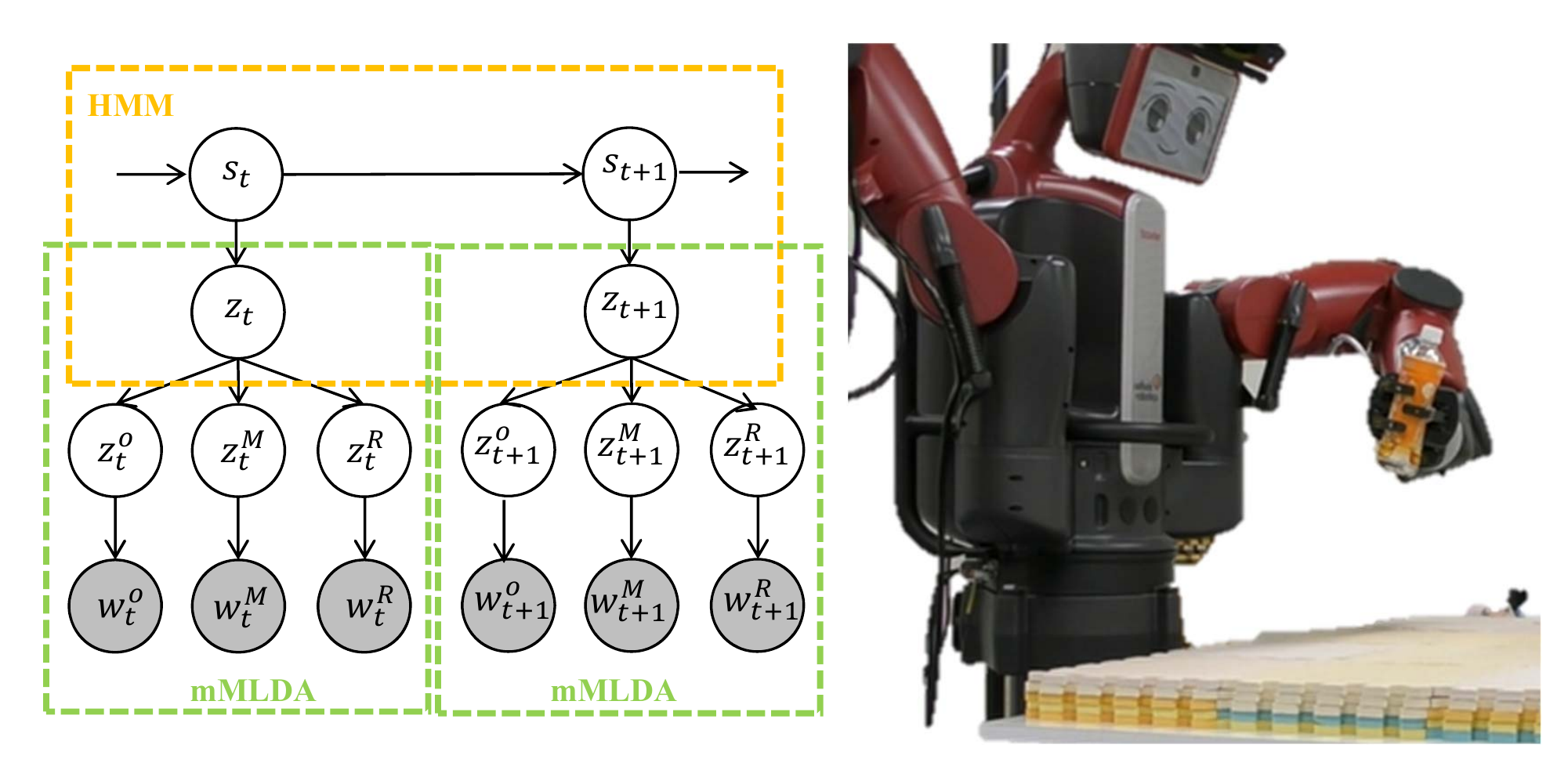}
 \caption{A robot planning and conducting a multimodal object categorization using a complex PGM~\cite{miyazawa2018}}
 \label{fig:baxter}
\end{figure}

This paper describes a novel framework enabling researchers and developers to create elemental cognitive modules, i.e., image recognition, automatic speech recognition, and syntax and clustering models, independently, and compose them into a large cognitive system, which can operate as a cognitive system and be consistently trained as a single learning system. 
Neuro-SERKET is an extension of SERKET~\cite{nakamura2017serket}, which was proposed as a framework for decomposing and composing PGMs.
As described later, SERKET does not support neural networks, i.e., deep learning.
A framework called Neuro-SERKET can also employ neural network-based cognitive modules.   In addition to that, SERKET only supports head-to-tail connections for decomposition and composition. In contrast, Neuro-SERKET supports head-to-head and tail-to-tail connections in graphical models, as well.

The remainder of this paper is organized as follows.
Section~\ref{sec:2} introduces the background of Neuro-SERKET. Section~\ref{sec:3} describes the Neuro-SERKET framework. More concretely, the method for the decomposition and composition of probabilistic generative models (PGMs), including DGMs, is described. Section~\ref{sec:4} describes a concrete example of integrative cognitive systems developed using the Neuro-SERKET framework. The integrative model was developed by combining VAE, a Gaussian mixture model (GMM), a latent Dirichlet allocation (LDA), and an automatic speech recognition system (ASR), and can form a multimodal concept from row speech and image signals. This is an illustrative example involving all types of elemental connections, i.e., head-to-tail, tail-to-tail, and head-to-head connections, and a neural network. Finally, Section~\ref{sec:5} provides some concluding remarks.

\begin{figure}[t]
 \centering
 \includegraphics[width=\linewidth]{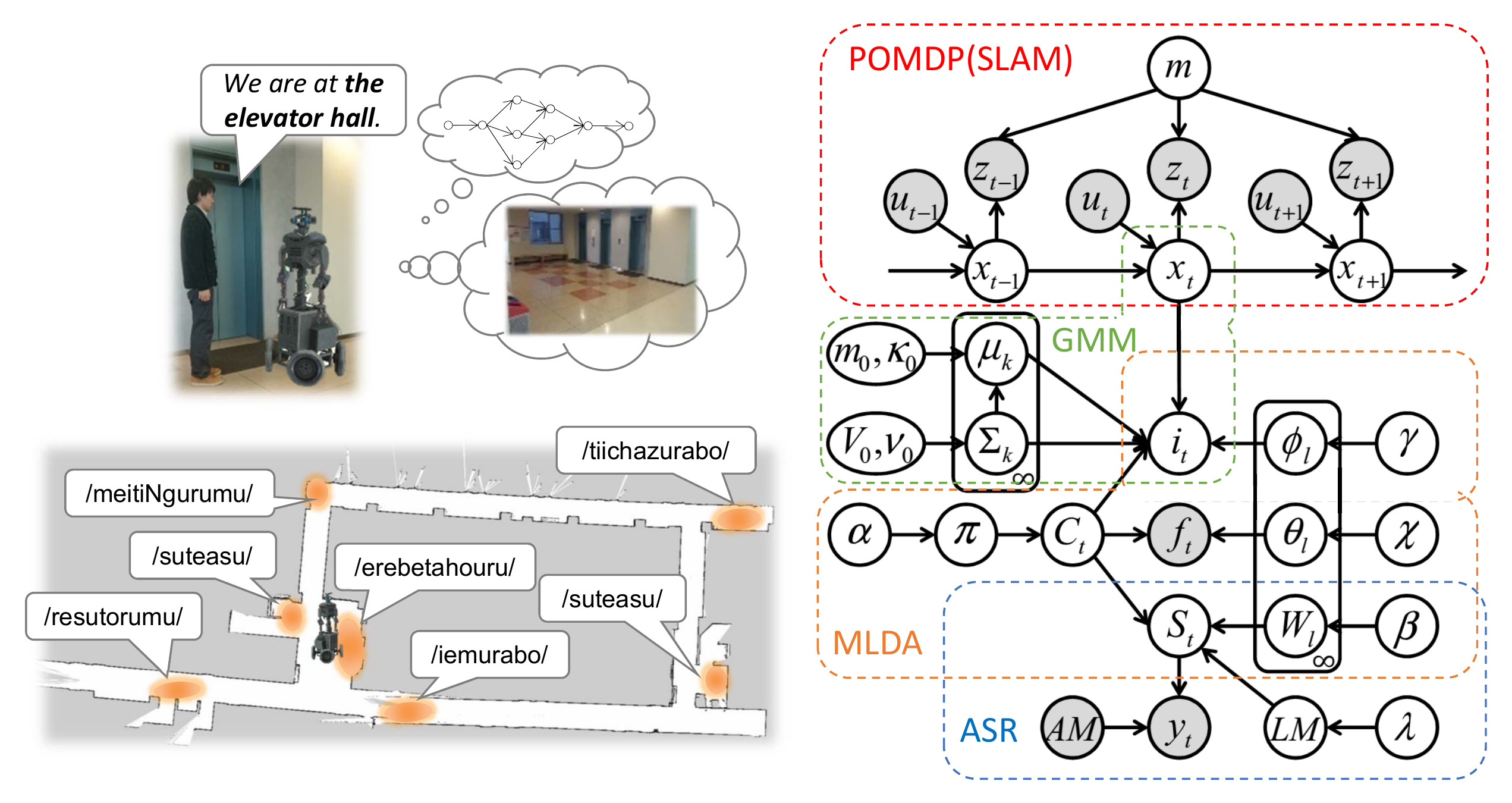}
 \caption{Graphical model of SpCoSLAM~\cite{taniguchi2017online}}
 \label{fig:spcoslam_pgm}
\end{figure}

\section{Backgrounds}\label{sec:2}
During this decade, the complexities of cognitive systems that can learn real-world knowledge and find the latent structure from multimodal sensorimotor information obtained by the robot itself, i.e., an embodied artificial cognitive system, have increased.
Cognitive systems for robots that learn the relationships among different types of multimodal sensory information have been proposed using PGMs and neural networks ~\cite{nakamuraIROS07,nishide2008,ogata2010inter,mangin2015mca,sinapov2014grounding}. 
Methods proposed in the studies enable robots to infer the latent variables from their own observations, for instance, the robot acquired object categories as latent variables from visual, sound and tactile sensory signals in~\cite{nakamuraIROS07}.
These enable robots to acquire various knowledge by inferring the latent variables from their own observations. 
A further advancement of such cognitive systems allows the robots to find meanings of words by treating a linguistic input as another modality \cite{roy:learning,nakamura_grounding,yamada2018paired}. 
Cognitive models have recently become more complex in realizing various cognitive capabilities: grammar acquisition \cite{attamimi2016learning}, language model learning \cite{joe2016}, hierarchical concept acquisition \cite{ando2013formation,hagiwara2018hierarchical}, spatial concept acquisition \cite{taniguchi2016spatial}, motion skills acquisition \cite{iwata2018learning}, and task planning \cite{miyazawa2018} (see Figure~\ref{fig:baxter}). It results in an increase in the development cost of each cognitive system.

Among them, it has been recognized that PGMs are extremely useful for modeling an integrative cognitive system that deals with multimodal and heterogeneous information and learns various functional concepts, i.e., internal representations, in an unsupervised manner because we can design the relationships of latent variables as a graphical model for introducing constraints to the data modeling. This can be interpreted through an analogy of designing cortical connections in our brain. Nakamura et al. proposed multimodal LDA (MLDA) for multimodal object categorization~\cite{nakamura_grounding}. They also developed a series of PGMs extending this idea. Taniguchi et al. proposed a spatial concept formation with simultaneous localization mapping (SpCoSLAM) for a spatial concept formation and lexical acquisition~\cite{taniguchi2017online} (see Figure~\ref{fig:spcoslam_pgm}). Such studies have contributed to the field of symbol emergence in robotics~\cite{Taniguchi2016symbol}.

A cognitive robot empowered by an integrative cognitive system can form object and spatial concepts, learn behaviors, and become able to understand human commands without explicit supervision differently from a supervised learning-based approach, which has been widely used in recent AI developments.
However, the growing complexity of graphical models has gradually increased barriers to entry into this research field for numerous researchers. A framework for developing an integrative cognitive system is required for further progress of this field in the same way as applied in accelerated studies on various deep learning frameworks around deep neural networks.

SERKET is a framework proposed for solving this problem~\cite{nakamura2017serket}. SERKET was designed to enable a distributed software development of extremely large PGMs.
In general, many pre-existing models for cognitive systems used in robots can be considered as a composition of elemental cognitive modules. For example, in Figures~\ref{fig:baxter} and \ref{fig:spcoslam_pgm}, the elemental modules in each graphical model are shown~\cite{miyazawa2018,taniguchi2017online}.
SERKET provides a theoretical framework for decomposing and composing PGMs. Cognitive modules developed in a distributive manner, namely, elemental PGMs, can be composed into a PGM using the SERKET framework, and the composed PGM can learn and work in the same manner as a PGM developed from scratch by a single developer.
However, SERKET has the following limitations.
\begin{itemize}
    \item SERKET only supports a head-to-tail connection, although in general, the graphical model can theoretically have tail-to-tail and head-to-head connections, as well.
    \item SERKET implicitly assumes the inference method using the Markov chain with a Monte Carlo approach and does not assume the integration of neural networks.  
\end{itemize}

The first limitation prevents us from a flexible, creative, and efficient development of a variety of integrative cognitive systems. For example, if we would like to develop an MLDA by integrating multiple LDAs, the framework should support tail-to-tail connections. 

The second limitation prevents us from the integration of DGMs, i.e., neural networks. As is widely known, DGMs can achieve representation learning, i.e., feature extraction. For example, a VAE is a probabilistic generative model having the capability of representation learning and can be integrated with PGMs, e.g., a hidden Markov model (HMM) and a GMM. However, SERKET does not support the integration of VAEs.

The integration of conventional PGMs, e.g., HMM and GMM, with DGMs, e.g., VAE, has received increasing attention, and such integrative PGMs have been studied. In a VAE, the encoder models the intractable posterior distribution of the latent representation, and the decoder reconstructs the observation using its latent representation, which usually assumes a single Gaussian prior. In recent studies, various PGMs such as GMM and HMM are applied to its latent space, and are used for semi-supervised learning\cite{kingma2014semi}, clustering\cite{johnson2016composing}\cite{dilokthanakul2016deep}, and acoustic unit discovery\cite{ebbers2017hidden}. The structured VAE (SVAE) proposed in \cite{johnson2016composing} is a generalization of the VAE to general PGMs, including capturing the correlation structure of sequential data, and in \cite{ebbers2017hidden}, it was extended to an acoustic unit discovery. In \cite{dilokthanakul2016deep}, a two-layer latent representation is composed that uniformly assumes a multi-modal prior distribution for a latent space, although this model requires a specific optimization to prevent an over-regularization. In\cite{jiang2016variational}, a generative process is defined based on a GMM in a latent space, and achieves a better performance. 

Not only the composition of a conventional PGM and a DGM, but also composition of DGMs should be explored.
More structured DGMs that can handle multimodal data are also gaining attention. Whereas vanilla VAEs can only take unimodal data, in\cite{sohn2015learning} and \cite{pandey2017variational}, conditional VAEs have been proposed that can handle another modality. These models can generate a modality corresponding to another modality data, e.g., generating images from captions~\cite{mansimov2015generating}. However, these models cannot generate multimodal data bi-directionally, i.e., both generating images from captions and generating captions from images, and also cannot obtain a representation that integrates their multimodal information. In \cite{suzuki2016joint}, a joint multimodal VAE is proposed, which not only has a multimodal inference model that embeds multimodal data into a joint representation but also unimodal inferences learned to approximate such multimodal data. The authors showed that, in the case of two modalities, this model can appropriately generate modalities bidirectionally and can infer a good joint representation. In \cite{wu2018multimodal}, the authors extended this multimodal inference model by introducing the idea of a product of experts~\cite{hinton2002training}, and proposed a multimodal VAE (MVAE) that can handle any number of modalities. Moreover,  \cite{jo2019cross} showed that a model whose association networks connect the latent variables of modality-specific VAEs can apply a cross-modal generation among multiple modalities. This line of studies clearly shows that DGMs become more and more structured and complex in the same way as conventional PGMs. Efficient way of developing complex cognitive systems by integrating DGMs should be explored.  

Considering the advancement of DGMs and recognizing the limitations of SERKET, in this study, we extend the application of SERKET and propose an updated version called Neuro-SERKET. Table~\ref{tbl:modules} shows a list of modules implemented in Neuro-SERKET library as examples. Developers can build a variety of integrative cognitive systems by composing these modules following the Neuro-SERKET framework.

\begin{table}[b]
\caption{Examples of modules implemented in current Neuro-SERKET library. }
\label{tbl:modules}
\centering
\begin{tabular}{|c|c|} \hline
Module	& Description	\\ \hline \hline
Observation			&	Module to deal with observations as messages \\ \hline 
CNNFeatureExtractor	&	Feature extractor from images based on CNN	\\ \hline
HACFeatureExtractor	&	Feature extractor from audio files based on HAC \cite{hac} \\ \hline
VAE					&	Module to learn feature representations based on VAE \cite{kingma2013auto} \\ \hline
MVAE				&	Module to learn features based on multinomial VAE \cite{Srivastava2017} \\ \hline
GMM				&	Unsupervised clustering based on GMM \\ \hline
MLDA				&	Unsupervised clustering based on MLDA \cite{nakamura_grounding} \\ \hline
MM					&	Module to learn transition of discrete variables \\ \hline
TtoT					&	Module to construct the tail to tail connection	\\ \hline
\end{tabular}
\end{table}

\section{Neuro-SERKET}\label{sec:3}
\subsection{Generation: Decomposition of Complex Graphical Model}\label{sec:2-1}

\begin{figure}[b]
 \centering
 \includegraphics[width=\linewidth]{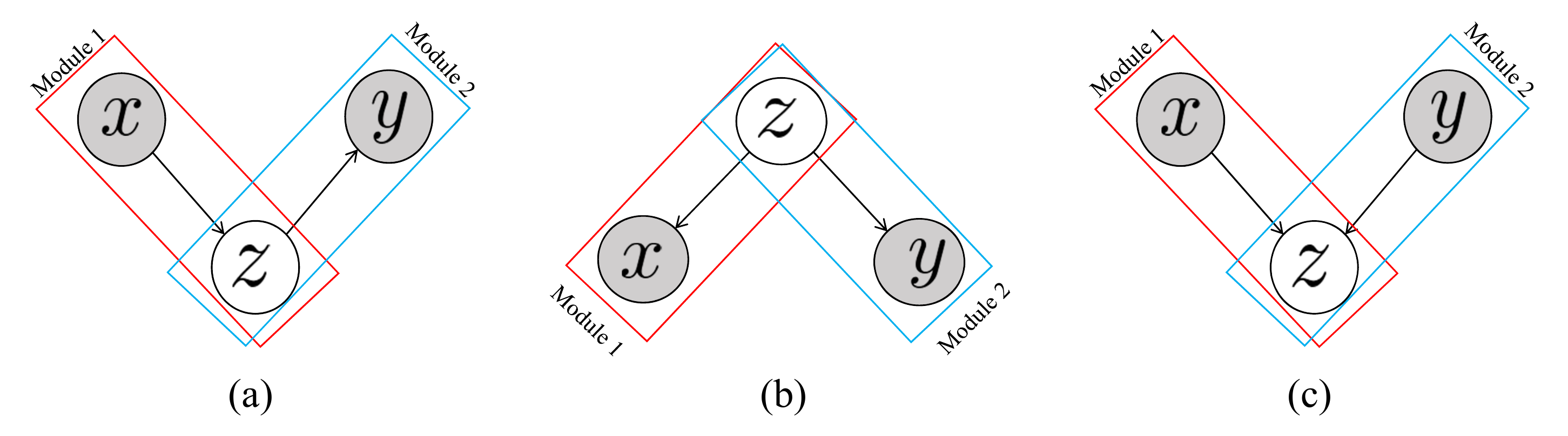}
 \caption{Elemental graphical models: (a) head-to-tail, (b) tail-to-tail, and (c) head-to-head}
 \label{fig:elemental_connections}
\end{figure}

\subsubsection{Overview}
Neuro-SERKET is an extension of SERKET. Therefore, it basically follows the approach of SERKET. 
SERKET provides a theoretical way to achieve a decomposition and composition of PGMs.
A decomposition is mainly related to the generative process, i.e., a generative model, and a composition is mainly related to the inference process, i.e., an inference model. 
In SERKET, decomposition and composition are conducted by following three rules.
\begin{enumerate}
\item A node, a latent variable $z$, in an integrated model is shared by two elemental modules. 
\item A module regards $z$ as an observable, and the parameter $\Theta$ of the probabilistic distribution $P(z|\Theta)$ is estimated.
\item The other module estimates $z$ by taking a prior $P(z|\Theta)$ with a fixed parameter $\Theta$, which is estimated in 2.
\end{enumerate}
Numerous types of PGMs can be described as graphical models. Directed graphs representing PGMs, i.e., graphical models, have three types of elemental connections, i.e., head-to-tail, head-to-head, and tail-to-tail (see Figure~\ref{fig:elemental_connections}).

The important feature of SERKET is that an integrated PGM developed by composing sub-modules following the SERKET framework can operate in almost the same way as a PGM developed from scratch, and uses an inference procedure developed for the PGM, in a reasonably approximate manner. Neuro-SERKET also has this feature.

In addition to a conventional SERKET framework, Neuro-SERKET provides two additional features.
\begin{itemize}
    \item Neuro-SERKET supports tail-to-tail and head-to-head connections in addition to head-to-tail connections.
    \item Neuro-SERKET supports deep probabilistic generative models, e.g., VAEs. Therefore PGMs using Neuro-SERKET can make use of the representation learning capability of neural networks.  
\end{itemize}

First, we describe how to decompose complex graphical models with the Neuro-SERKET framework.
As is widely known, probabilistic graphical models have three types of elemental connections, as shown in Figure~\ref{fig:elemental_connections}. Note that each generative process, e.g., $P(x|z)$, has global parameters, e.g., $\theta$ for $P(x|z, \theta)$, although these are omitted from the graphical model for the sake of simplicity. Each generative process can have other latent variables as well. A systematic approach to a decomposition is also described herein.

\subsubsection{Head-to-tail decomposition}

In the Neuro-SERKET framework, a complex graphical model is systematically decomposed.
First, we take a head-to-tail connection, shown in Figure~\ref{fig:elemental_connections}~(a), as an example.
The joint distribution $P(x,y,z)$ can be written as follows because of a conditional dependency indicated by the graphical model.
\begin{eqnarray}
P(x,y,z) = P(y|z) P(z|x) P(x) 
\label{eq:chain_joint}
\end{eqnarray}
The generative process of the latent variable $z$ is described as $P(x, z)=P(z|x)P(x)$. 
Next, when looking at the generative process of $y$, it can be seen that the generative process of $y$ can be described as $P(y, z|x=X)=P(y|z) P(z|x = X)$, where $X$ is an instance of $x$. Here, note that $P(y, z|x=X)$ does not depend on the variable $x$ when $x$ is fixed, i.e., $x=X$. 
This means the probabilistic generative model can be decomposed into two modules.

The discussion above is reconfirmed from the viewpoint of factorization. The joint probability can be factorized in two ways. 

\begin{eqnarray}
P(x,y,z) &=& P(y | z)\underbrace{P(z, x)}_{\mbox{Module 1}} \label{eq:module1}\\
&=& \underbrace{P(y, z|x)}_{\mbox{Module 2}} P(x)\label{eq:module2}
\end{eqnarray}
The first and second modules correspond to the generative model for $z$ and $y$, respectively.

If a joint distribution can be factorized in two ways when sharing a latent variable, e.g., $z$ in Equations \ref{eq:module1} and \ref{eq:module2}, the PGM can be decomposed into two modules.

We introduce an operator $\otimes$ representing a composition operation of PGMs for illustrative purposes. 

\begin{eqnarray}
P(x,y,z) \Rightarrow  P(z, x) \otimes P(y ,z)
\end{eqnarray}
This shows that PGM $P(x,y,z) $ can be decomposed into $P(z, x)$ and $P(y ,z)$, which are two elemental modules.

\subsubsection{Tail-to-tail decomposition}

Another elemental connection is a tail-to-tail (see Figure~\ref{fig:elemental_connections}~(b)) connection, which is also called a ``fork.''
The joint distribution of $x, y$, and $z$ can be described as follows using an assumed conditional independence.
\begin{eqnarray}
P(x,y,z) = P(x|z) P(y|z) P(z) 
\label{eq: fork_joint}
\end{eqnarray}
In the same way as the discussion regarding a head-to-tail connection, the joint distribution can be factorized in the following two ways.
\begin{eqnarray}
P(x,y,z) &=& \underbrace{P(x,z)}_{\mbox{Module 1}} P(y|z)\\
&=& P(x|z) \underbrace{P(y,z)}_{\mbox{Module 2}}
\end{eqnarray}
Each module obtained through a decomposition corresponds to a generative process of $x$ and $y$.
Following the usage of symbol $\otimes$, which we introduced in the previous subsection, the PGM $P(x,y,z)$ can be decomposed into two modules, i.e., joint distributions, $P(x, z)$ and $P(y, z)$, as follows. 

\begin{eqnarray}
P(x,y,z) \Rightarrow P(x ,z) \otimes P(y, z)
\end{eqnarray}

\subsubsection{Head-to-head decomposition}
The other elemental connection is a head-to-head (see Figure~\ref{fig:elemental_connections}~(c)) connection.
The joint distribution of $x, y$, and $z$ under a head-to-head connection can be decomposed when considering the following conditional independence. 
\begin{eqnarray}
P(x,y,z) = P(z|x,y) P(x) P(y) 
\label{eq: collider_joint}
\end{eqnarray}
If we apply the systematic rule for a decomposition in the same way as a head-to-tail and tail-to-tail connection, we will obtain the following decomposition.
\begin{eqnarray}
P(x,y,z) &=& \underbrace{P(x,z|y) }_{\mbox{Module 1}}P(y)\\
&=& \underbrace{P(y, z|x) }_{\mbox{Module 2}} P(x)
\end{eqnarray}
However, differing from the previous decomposition, i.e., head-to-tail and tail-to-tail connections, both modules represent the generative process of $z$, and involve $x$ and $y$. This prevents us from taking a SERKET-based approach for inferring $z$ in each module because both of the modules involve $x$ and $y$.
The SERKET framework requires that a latent variable $z$ be inferred within a module using one of $x$ or $y$ after decomposition. In other words, $z$ should be regarded as an observable, i.e., a given variable, in another module. Therefore, SERKET does not provide the way of decomposition for a head-to-head connection.
However, the decomposition of a head-to-head connection is important in building further complex cognitive systems. For example, SpCoSLAM assumes that a generated sentence $S_t$ is conditioned by the spatial concept $C_t$, i.e., ``where the robot is,'' and syntactic and lexical information, i.e., a language model $LM$ and the set of parameters of topic-dependent word distributions $\{W_l\}$ (see Figrue~\ref{fig:spcoslam_pgm})~\cite{taniguchi2017online}.

Therefore, in Nuero-SERKET, we introduce a new way to achieve an approximate decomposition for $P(z|x,y)$.

\begin{eqnarray}
P(z|x,y) \approx \hat{P}(z|x,y) \propto P(z|x) P(z|y), 
\label{eq: collider_joint_approx}
\end{eqnarray}
where $\hat{P}(z|x,y)$ is an approximately decomposed distribution. This approximation consists of two steps. The first approximation is that $P(z|x,y)$ is decomposed into distributions including $P(z|x)$, $P(z|y)$ and $P(z)$. This approximate decomposition can have two ways of interpretation: a product of expert (PoE), i.e., 
$\hat{P}(z|x,y) = P(z)P(z|x)P(z|y)$ ~\cite{hinton2002training}, 
and a uni-gram re-scaling, i.e., 
$\hat{P}(z|x,y) =\frac{P(z|x)P(z|y)}{P(z)}$~\cite{gildea1999topic}. 
In both cases, a prior $P(z)$ is considered to be a uniform distribution. This means that the prior in the distribution $\hat{P}$ can be ignored, i.e., $\hat{P}(z|x,y) \propto P(z|x) P(z|y)$.

% Two methods of interpretation can be applied to this approximation.
% The first interpretation is a product of expert (PoE)~\cite{poe_hinton}. In particular, in the context of variational inference, this type of approximation is widely used. The other is a two-step approximation using a uni-gram re-scaling~\cite{gildea1999topic} and uniform distribution approximation of a prior $P(z)$\footnote{\
% The two-step approximation is as follows:
% \begin{eqnarray}
% P(z|x,y) &\approx& P^{UR}(z|x,y) \propto 
% \frac{P(z|x) P(z|y) }{P(z)}  \qquad \text{(Uni-gram re-scaling)} \nonumber\\
% &\approx& \hat{P}(z|x,y) \propto P(z|x) P(z|y)  \qquad \text{(Uniform distribution approximation of $P(z)$)}\nonumber
% \label{eq: collider_joint_twostep_approx}
% \end{eqnarray}
% }.

Using this approximation, we can obtain the following modules.
\begin{eqnarray}
P(x,y,z) &=& P(z|x,y) P(x) P(y) \\
&\approx\propto&P(z|x)P(z|y) P(x) P(y) \\
&=& \underbrace{P(x,z)}_{\mbox{Module 1}}\underbrace{P(y,z)}_{\mbox{Module 2}}
\label{eq: collider_joint_decomposition}
\end{eqnarray}
where $\approx\propto$ represents the terms ``approximation'' and ``proportion to''\footnote{$P(x) \approx\propto f(x)$ is an abbreviation of $P(x) \approx \hat{P}(x) \propto f(x)$. }.  

The decomposition is described as follows.
\begin{eqnarray}
P(x,y,z) \Rightarrow P(x ,z) \otimes P(y, z)
\end{eqnarray}

The appropriateness of the approximation is evaluated empirically in Section~\ref{sec:4} based on an experiment.

For example, SpCoSLAM (Figure~\ref{fig:spcoslam_pgm}) has a head-to-head connection at approximately $S_t$, i.e., a sentence recognized by an ASR system. When we pick up related variables for illustrative purposes, we can start with the following joint distribution.

\begin{eqnarray}
P(y_t, LM, S_t, C_t | AM) = P(y_t|AM, S_t) P(S_t|C_t, LM) P(LM) P(C_t)
\end{eqnarray}
where $y_t$ and $AM$ are a speech signal and acoustic model in an ASR system, respectively. 

In practical terms, $AM$ and $LM$ are implemented in an ASR system, i.e., packaged software, and $C_t$ is a part of a multimodal categorization module. Therefore, calculating a generative probability and drawing samples theoretically are extremely difficult. Therefore, Neuro-SERKET introduces the following approximation\footnote{Note that the original study on SpCoSLAM does not use this method of approximation. A uni-gram rescaling approximation alone was employed instead.}.

\begin{eqnarray}
% P(S_t| C_t, LM ) \approx \hat{P}(S_t| C_t, LM ) \propto P(S_t| LM ) P(S_t| C_t )
P(S_t| C_t, LM ) \approx\propto P(S_t| LM ) P(S_t| C_t )
\end{eqnarray}

This allows developing a graphical model with two parts, i.e., an ASR (including a lexical acquisition) module and a multimodal categorization module\footnote{Note that, for illustrative purposes, the other variables and hyperparameters are ignored from the equations.}. 

\begin{eqnarray}
P(y_t, LM, S_t, C_t| AM) \approx\propto \underbrace{ P(y|AM, S_t) P(S_t| LM) P(LM) }_{\mbox{ASR module}} \underbrace{P(S_t|C_t) P(C_t)}_{\shortstack{Multimodal\\ categorization\\ module}}
\end{eqnarray}

\subsection{Inference: Composition of Complex Graphical Model}\label{sec:2-2}
In the SERKET framework, each module is developed by a different researcher or developer in a distributed manner. After the development of the modules, they are integrated into an integrative cognitive system.
Integrated modules learn together and work together as a single cognitive system.

\begin{figure}
\centering
 \includegraphics[width=\linewidth]{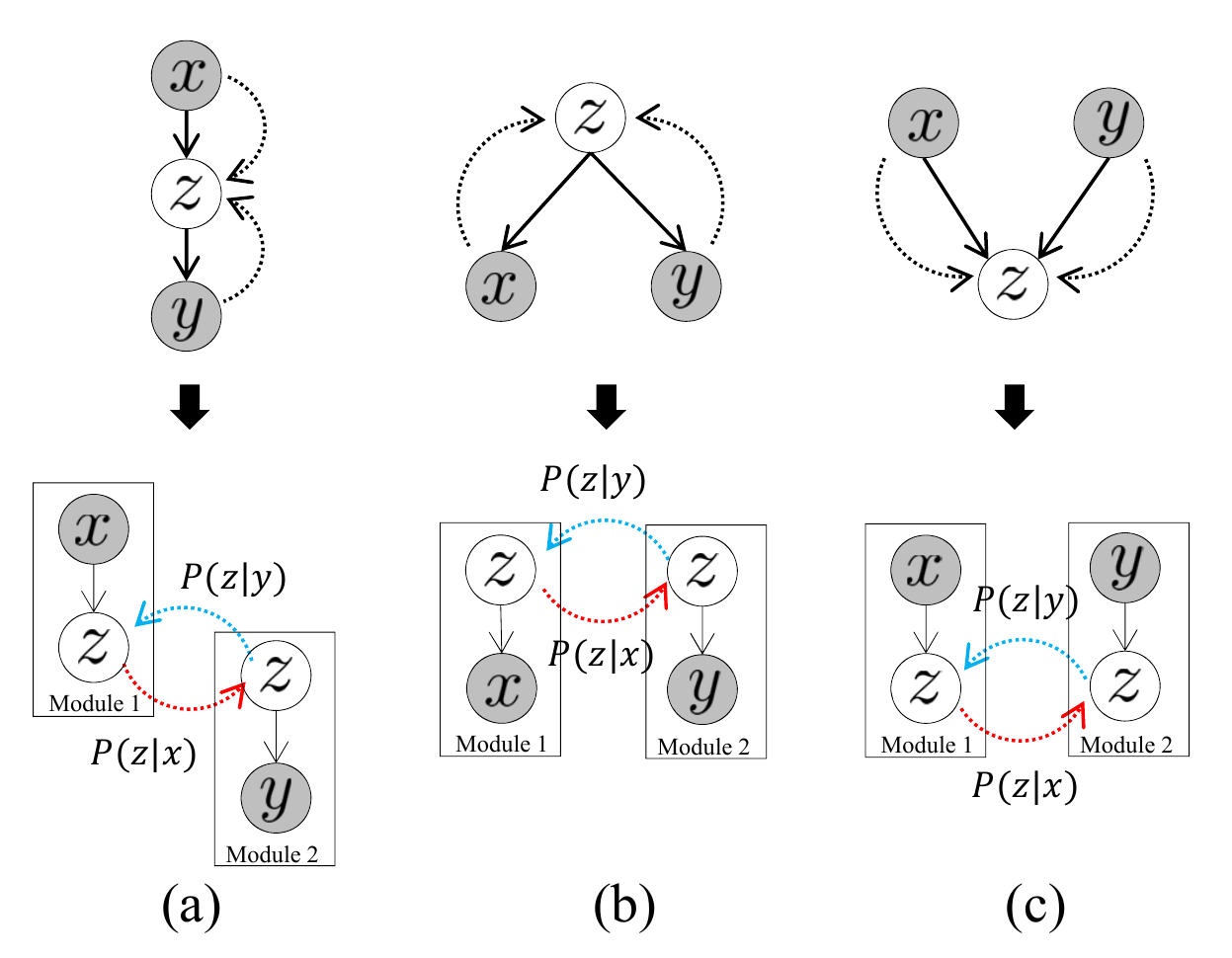}
\caption{Graphical models and their Neuro-SERKET implementations of (a) head-to-tail, (b) tail-to-tail, and (c) head-to-head. The black-dotted arrows represent conditional probabilities used in the inference. }
\label{fig:inference}  
\end{figure}

In the context of PGMs, prediction, estimation, and learning are simply regarded as an inference of latent variables. Therefore, the composition of the modules corresponds to an inference procedure combining multiple modules. This section provides a method of composition for each elemental connection. Figure~\ref{fig:inference} summarizes the three types of elemental connections, message passing, and the decomposition method used in our framework, Neuro-SERKET.
In each graphical model, the black-dotted arrows indicate the calculation of a posterior distribution, which is necessary for an inference procedure.
Note that the two dotted arrows in each graphical model form head-to-head relationships.

Figure~\ref{fig:inference} (a) shows the method of message passing between two modules in the case of a head-to-tail connection. 
Conventional SERKET mainly introduced two methods for achieving a head-to-tail composition.

The first is called the message passing (MP) approach, and its procedure is as follows\footnote{As a variation to the MP approach, module 1 can send samples, i.e., the data distribution, $z^l \sim P(z|x) (l =\{1, \ldots, L\})$, as a Monte Carlo approximation of $P(z|x)$. As a special case of this, module 1 can send a sample $z^\star \sim P(z|x)$ to module 2. In Section \ref{sec:4}, as an example, the VAE module sends a recognition result to another module.}.
\begin{enumerate}
    \item In module 1, $P(z|x)$ is computed.
    \item $P(z|x)$ is sent to module 2.
    \item In module 2, the probability distribution $P(z|y)$, which represents the relationships between $x$ and $y$, is estimated using $P(z|x)$.
    \item $P(z|y)$ is sent to module 1.
    \item In module 1, the latent variable $z$ is estimated, and the parameters of $P(x|z)$ are updated.
\end{enumerate}
%\comment{tanichu}{VAE対応の仮の説明}

The other is called a sampling importance resampling (SIR) approach, the procedure of which is as follows.
\begin{enumerate}
    \item Generate $L$ samples $z^{(l)} \sim P(z|x)$ in module 1.
    \item Send $\{ z^{(1)}, \ldots ,z^{(L)}\}$ to module 2.
    \item Select sample $z^{\star}$ among $\{ z^{(1)}, \ldots ,z^{(L)}\}$ by calculating their importance using $P(z|y)$ and update the parameters of $P(z|y)$. 
    \item Send the selected sample $z^{\star}$ to module 1.
    \item Update the parameters of $P(x|z)$.
\end{enumerate}
This approach involves a Monte Carlo approximation. However, many off-the-shelf modules do not support the calculation of a posterior distribution $P(z|x)$ itself. Therefore, the SIR approach allow us to use various off-the-shelf modules, e.g., ASR and image recognition systems, that provide only samples, i.e., estimated results.

With this inference procedure, SERKET employs a PoE approximation, i.e., $P(z|x,y)\approx\propto P(z|x)P(z|y)$ in the same way as a head-to-head decomposition. For further details, please refer to the original study~\cite{nakamura2017serket}.

Differing from the decomposition part, there are no structural differences among the three elemental connections. 
The dotted line in Figure~\ref{fig:inference} shows the inference process for each elemental connection. 
We can see that all pairs of dotted arrows have head-to-head connections. 
This means that, in the inference process, i.e., composition, we can use the same procedure as a head-to-tail composition in the cases of tail-to-tail and head-to-head compositions.

However, we need to develop a special treatment for the implementation of a tail-to-tail composition. SERKET requires connecting latent variables of modules in a hierarchical manner~\cite{nakamura2017serket}, and we cannot connect module 1, i.e., $P(x|z)$, and module 2, i.e., $P(y|z)$, directly in a tail-to-tail composition (Figure~\ref{fig:inference} (c)). Therefore, we introduce an auxiliary module called a tail-to-tail (TtoT) module, which connects modules 1 and 2 and transfers the probability distribution between the two modules.  

In this way, Neuro-SERKET also makes use of a PoE and uniform distribution prior approximation\cite{hinton2002training} in the composition, and achieves an inference of the integrative PGMs. 
Differing from these assumptions, description, and discussion in the original study on SERKET~\cite{nakamura2017serket}, Neuro-SERKET does not assume any specifics for the implementation and inference procedure of each module\footnote{In the original study on SERKET, the authors mentioned that they ``employed a
sampling-based method because of its simpler implementation.'' This means that they excluded modules that are trained using other inference procedures, e.g., gradient-based methods. In general, a sampling-based approach is unsuitable for the training of neural networks. This means that SERKET fails to involve neural network-based modules, which have recently been widely used, into SERKET-based cognitive systems.}. Therefore, Neuro-SERKET can integrate neural network-based generative models, i.e., DGMs such as VAEs.

\section{Example: Concept Formation using VAE+GMM+LDA+ASR}\label{sec:4}
This section describes an illustrative example of a cognitive system that can be developed following the Neuro-SERKET framework by integrating pre-existing modules.
For illustrative purposes, this model involves all of the elemental connections, i.e., head-to-tail, tail-to-tail, and head-to-head. A neural network-based module, i.e., VAE, is also included as an elemental module.
The developed PGM for multimodal categorization is a composition of a VAE, GMM, LDA~\cite{Blei:2003:LDA:944919.944937}, and ASR. 
We empirically validated Neuro-SERKET through an experiment using image data and speech signals.

\subsection{Model}
\begin{figure}
\centering
 \includegraphics[width=0.8\linewidth]{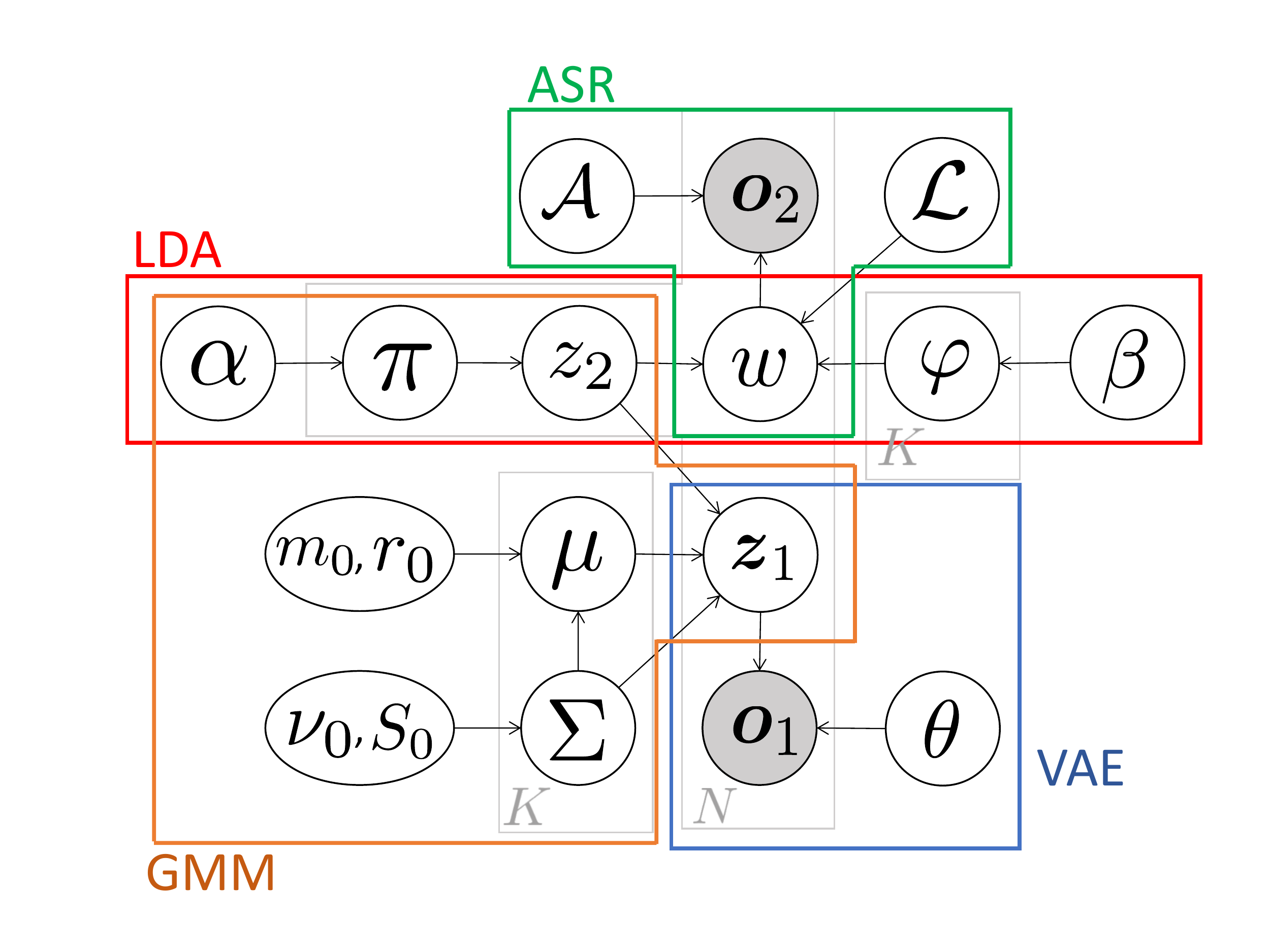}
\caption{The original graphical model of the integrative PGM (VAE+GMM+LDA+ASR). Each block shows each module.}
\label{fig:graphicalodel}       % Give a unique label
\end{figure}

Figure~\ref{fig:graphicalodel} shows a graphical model of the PGM developed using the Neuro-SERKET framework. 
This PGM receives two types of observations, i.e., pairs of an image $\bo_1$ and speech signal $\bo_2$ corresponding to the image. The PGM is for an unsupervised multimodal categorization, including representation learning of the image data.
Image $\bo_1$ is expected to be encoded into the latent variable $\bz_1$ using VAE.
The speech signal $\bo_2$ is recognized, and word $w$ is estimated using a language model parameterized by $\cal L$, which can also be learned from the data.
The obtained word $w$ is clustered using LDA, and the estimated representation of image $\bz_1$ is clustered using GMM.
Note that the latent variable $\bz_2$ representing the class of the input pair of data is shared by the LDA and GMM. 
This means that an estimation of $\bz_2$ corresponds to a multimodal categorization. 
A list of parameters of the PGM is enumerated in Table~\ref{tbl:model_param}.

\begin{table}[b]
\centering
\caption{Model parameters of the integrative PGM (VAE+GMM+LDA+ASR). }
\label{tbl:model_param}
\begin{tabular}{|c|c|} \hline
Parameter & Description \\ \hline
$\theta$ & Parameter of VAE decoder (generative network)\\ \hline
$\bo_1, \bo_2$ & Observations, image data and speech signal\\ \hline
$\bz_1$ & Latent variable of VAE extracted from $\bo_1$ \\ \hline
$\bz_2$ & Index of classes the observations are categorized into\\ \hline
$ w$ & Word recognized by the ASR system\\ \hline
$\bmu, \bSigma$ & Mean vector and variance-covariance matrix of Gaussian distribution \\ \hline
$r_0, m_0, S_0, \nu_0$ & Parameters of Gauss-Wishart distribution \\ \hline
$\pi, \varphi$ & Parameters of multinomial distribution \\ \hline
$\alpha, \beta$ & Parameters of Dirichlet distribution \\ \hline
$N$ & The number of observations \\ \hline
$K$ & The number of classes in LDA and GMM \\ \hline
 \end{tabular}
\end{table}

\begin{figure}
\centering
 \includegraphics[width=1.0\linewidth]{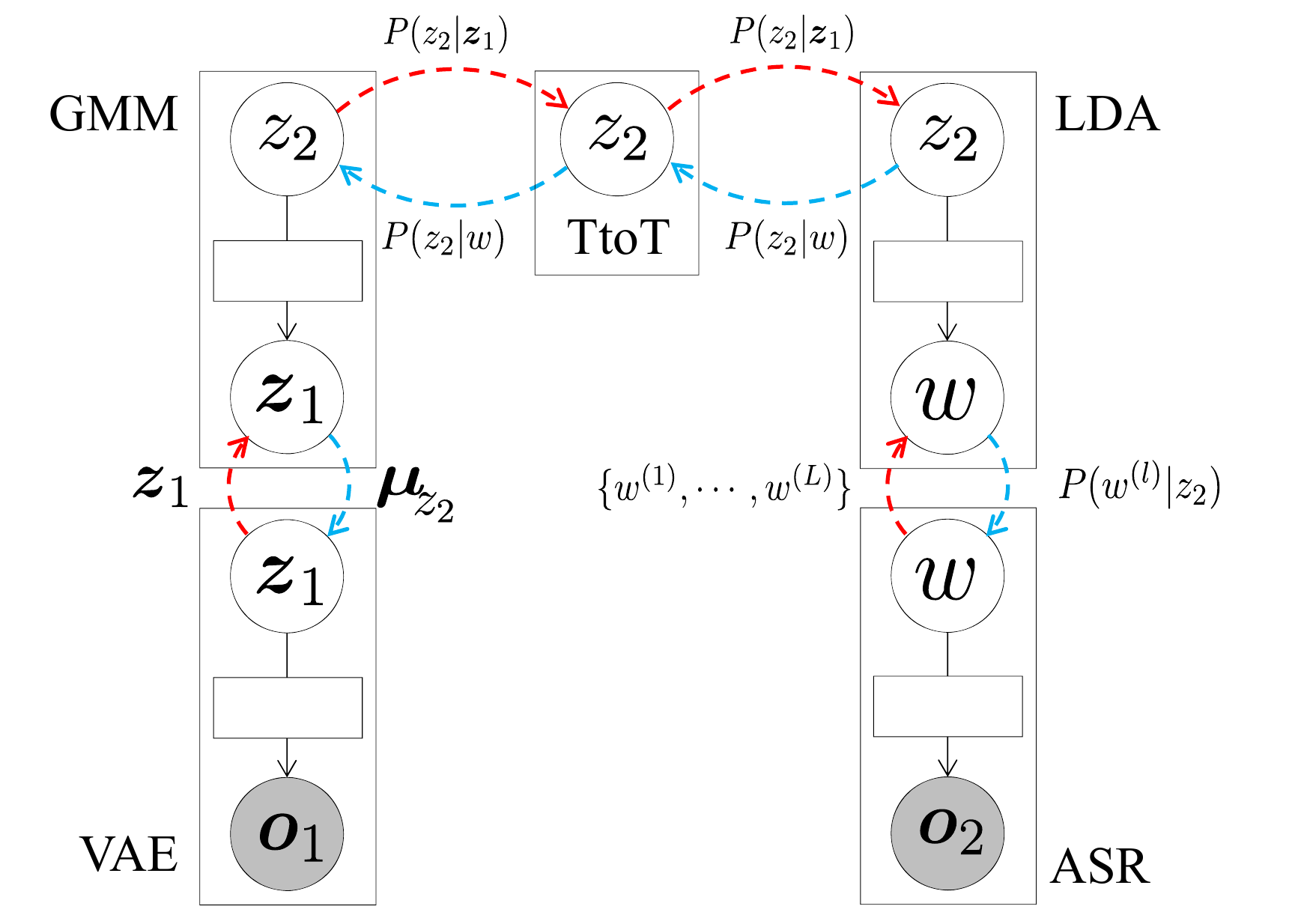}
\caption{Decomposed modules and communication between them following the Neuro-SERKET framework.}
\label{fig:serket_model}       % Give a unique label
\end{figure}

Figure~\ref{fig:serket_model} shows the elemental cognitive modules and communication between them. The communication conducted during the inference procedure, i.e., the composition, is summarized as follows.

\begin{description}
   \item[\bf VAE module] The VAE module extracts a representation, i.e., latent variable, $\bz_1$, from the image data $\bo_1$, and sends $\bz_1$ to the GMM module. The GMM module sends $\bmu_{z_2}$, which is a mean vector of the Gaussian distribution that $\bz_1$ was categorized into, back to the VAE module.
   VAE uses $\bmu_{z_2}$ and updates the parameters of the encoder and decoder of the VAE to maximize the evidence lower bound (ELBO).  %
   \begin{eqnarray}
   {\cal L}(\btheta, \bphi; \bo_1) = -D ( q_\phi (\bz_1 | \bo_1) || {\cal N} (\bmu_{z_2}, \bI ) ) + E_{q_\phi (\bz_1|\bo_1)} [ \log p_\theta ( \bo_1 | \bz_1 ) ]
   \end{eqnarray}
   This allows the VAE to learn a representation appropriate for categorization by the GMM module.
   
   \item[\bf GMM module] The GMM module sends $P(\bz_2| \bz_1)$, which is obtained by categorizing $\bz_1$ received from the VAE module, to the TtoT module . 
   This module shares $\bz_2$ with the LDA module (Figure~\ref{fig:graphicalodel}).
   Therefore, the inference of $bz_2$ is affected by the LDA module. 
   The TtoT module mediates the information between the GMM and LDA modules.  
   When the GMM module applies an inference, i.e., a categorization, the GMM module uses $P(\bz_2|\bw)$, which is received from the TtoT module.
   \begin{eqnarray}
   \bz_2 \sim P(\bz_2|\bz_1 , \bw)  \approx\propto P(\bz_2 | \bz_1) P (\bz_2 | \bw) 
   \end{eqnarray}

   \item[\bf ASR module] The ASR module represents an off-the-shelf speech recognition system\footnote{Julius: Open-Source Large Vocabulary Continuous Speech Recognition Engine: \url{https://github.com/julius-speech/julius}}.
   The ASR module sends the $L$-best speech recognition results of $\bo_2$ to the LDA module. The $L$-best results are regarded as an approximation of $L$ samples from the posterior distribution $P(w|\bo_2)$.
   The LDA module calculates the importance of each word $P(w^{(l)}|\bz_2)$, and re-samples the word using the importance weight (SIR approach) as follows. 
   \begin{eqnarray}
   w^{(l)} &\sim& P(w|\bo_2) \\
  % w^* \sim P(w^{(l)} | \bz_2)
   w^* &\sim& \hat{P}(w) \propto \sum_l P(w^{(l)} | \bz_2) \delta_{w^{(l)}}(w),
   \end{eqnarray}
   where $\delta_{w^{(l)}}(w)$ is a probability mass function. 

   \item[\bf LDA module] The LDA module receives a set of words $\bw= \{ w^{(1)}, \cdots, w^{(L)} \}$ and clusters them. As a result, the LDA module calculates $P(\bz_2|\bw)$ and sends it to the TtoT module. 
   Note that $\bz_2$ is shared with the GMM module (see Figure~\ref{fig:graphicalodel}), and in the clustering, i.e., inference, the process is affected by the categorization by the GMM module. 
   Therefore, the LDA module uses $P(\bz_2|\bz_1)$ received from the TtoT module when it clusters words.
   \begin{eqnarray}
   \bz_2 \sim P(\bz_2 | \bz_1) P (\bz_2 | \bw) 
   \end{eqnarray}

   \item[\bf TtoT module]%
    The TtoT module simply transfers $P(\bz_2|\bw)$ from the LDA module to the GMM module, and $P(\bz_2|\bz_1)$ from the GMM module to the LDA module.
\end{description}
Following the communication procedure shown above, the total PGM can be trained by optimizing each module locally under the influence of neighboring modules.
  
\subsection{Code}
\begin{figure}[h]
\centering
\includegraphics[width=7.0cm]{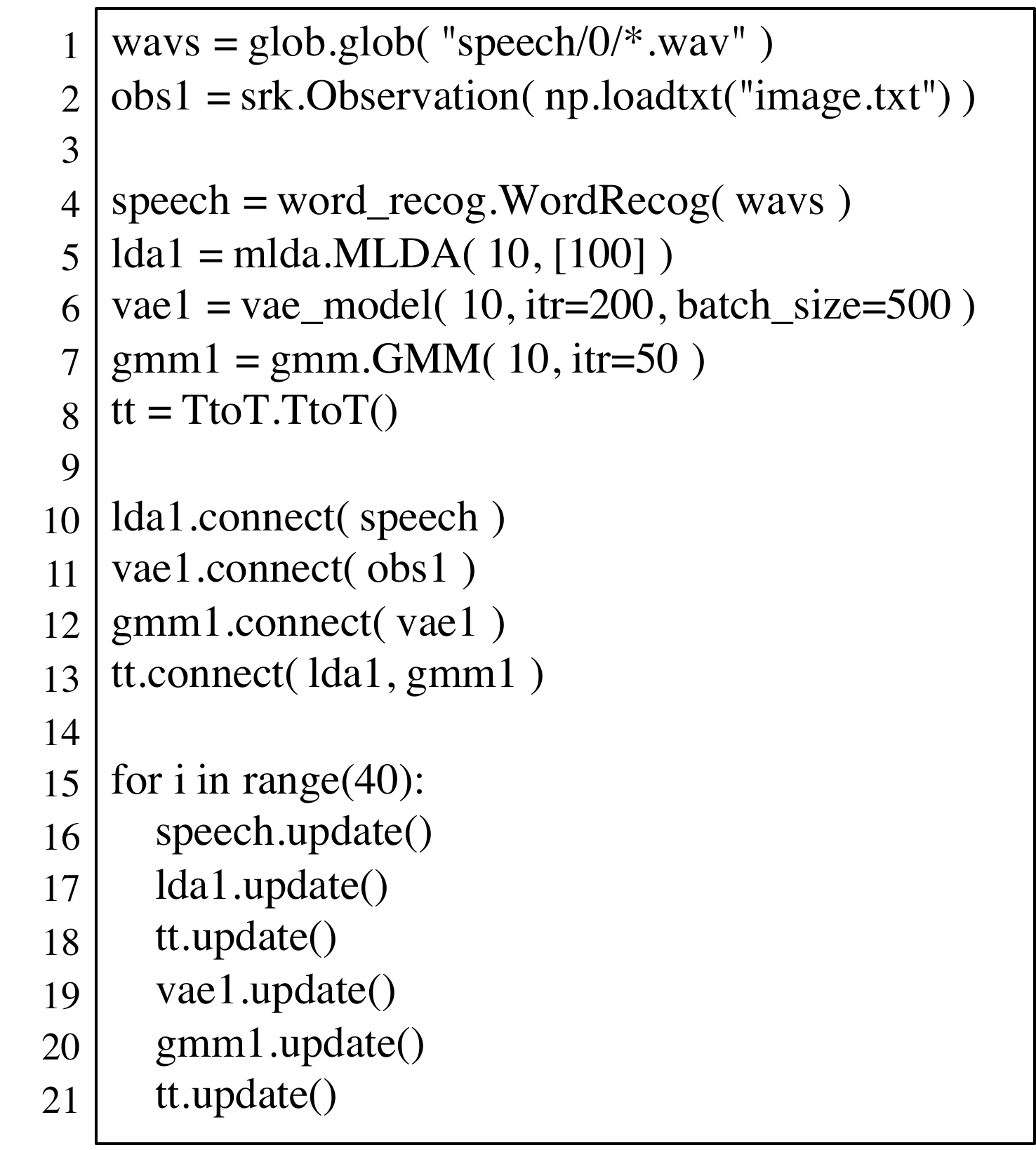}
\caption{Source code of main part of the implementation example. }
\label{fig:code}
\end{figure}
Figure~\ref{fig:code} shows the source code of the main part of the implementation. 
The observations are loaded from files in lines 1-2, the modules to be used are defined in lines 4-8, the connections between the modules are defined in lines 10-13, and the parameters are estimated in lines 15-21. 
The total number of lines is less than 80, including other parts such as import syntax and the definition of the structure of VAE. 
Note that even more complicated models can also be implemented in a few hundreds of lines, and we believe the current Neuro-SERKET has scalability with regard to programmatic implementation.

\subsection{Conditions}

\begin{table}
\caption{Pronunciation of Japanese numbers used in the experiment. }
\label{tbl:japanese_number}
\centering
\begin{tabular}{lclcl} \hline
Number & Japanese pronunciation  \\ \hline
0 & ze ro \\ \hline
1  & i chi \\ \hline
2 & ni \\ \hline
3 & sa n \\ \hline
4 & yo n \\ \hline
5 & go \\ \hline
6 & ro ku \\ \hline
7 & na na \\ \hline 
8 & ha chi \\ \hline
9 & kyu u \\ \hline
\end{tabular}
\end{table}
During the experiment, we used a hand-written digit dataset, MNIST\cite{lecun2010mnist}, and a spoken Japanese number dataset~\cite{censrec4}, as the image data and speech signals, respectively. Each pair of data consist of image data and a spoken audio signal corresponding to a number among $\{ 0, \ldots, 9\}$.  In total, 3,000 pairs are used.
The pronunciation of Japanese digits is shown in Table~\ref{tbl:japanese_number}. 

We used VAE, whose encoder and decoder have a middle layer with 128 nodes and a hidden layer with ten nodes, i.e., the dimension of the latent space was $10$. The number of classes of GMM and LDA was $K=10$. We used Julius for the ASR module. We used a standard GMM-based acoustic model preset in Julius, and a language model in which all syllables have the same probability as an initial language model.
The number of samples obtained from the ASR module was $L=10$. 
%The parameters of all modules were updated 50 times.

During the experiment, we compared the following four models.
\begin{description}
  \item[\bf VAE GMM LDA ASR] No communication among the modules
  \item[\bf VAE GMM LDA+ASR] Communication between LDA and ASR 
  \item[\bf VAE+GMM LDA+ASR] Communication between VAE and GMM and between LDA and ASR
  \item[\bf VAE+GMM+LDA+ASR] Communication among all modules
\end{description}
In the name of each model, `+' represents the existence of communication between the two modules, and ` '(white space) indicates no communication between the two neighboring modules. Note that none of the three connections were not supported in SERKET framework.

In the learning process, the model is trained in an off-line manner. 
Posterior probabilities for all data points are calculated and were given to the neighbor modules. When a module is updated, the global parameters of the module was reset once and trained using the received data and messages. In each update, VAE was trained 200 epochs with batch size $=500$, and  GMM and LDA were trained with Gibbs sampling procedure with $50$ and $100$ times sampling, respectively.
VAE+GMM, LDA+ASR, and VAE+GMM+LDA+ASR were updated 50 times, i.e., until they converged.  
The order of the update were {\bf ASR}$\rightarrow$ {\bf LDA} (from LDA to GMM)$\rightarrow$ {\bf TtoT}$\rightarrow$ {\bf VAE}$\rightarrow$ {\bf GMM}$\rightarrow$ {\bf TtoT} (from LDA to GMM)$\rightarrow$ {\bf ASR}.

The average of accuracy was calculated by referring to the ground-truth category of the digits for each condition.

\subsection{Result}

\begin{table}
\caption{Classification accuracy in the GMM and LDA modules. }
\label{tbl:ari}
\centering
\begin{tabular}{lccccc} \hline
model & \multicolumn{2}{c}{Accuracy (\%)} & \multicolumn{3}{c}{Features introduced in Neuro-SERKET}\\ \hline
                & GMM & LDA &Head-to-head&Tail-to-tail&Neural net\\ \hline
VAE GMM LDA ASR & 62.0    & 27.4  &&&    \\ \hline
VAE GMM LDA+ASR & 62.0   & 91.8 &\checkmark & &     \\ \hline
VAE+GMM LDA+ASR & 63.7  & 91.8  &\checkmark& &\checkmark    \\ \hline
VAE+GMM+LDA+ASR & {\bf 91.0}  &  {\bf 93.7}  &\checkmark&\checkmark&\checkmark   \\ \hline
\end{tabular}
\end{table}

Table~\ref{tbl:ari} shows the average level of accuracy for each clustering module, i.e., GMM and LDA modules.

The performance of the GMM module is slightly increased by introducing communication with a VAE.
In contrast, the performance of the LDA module is significantly increased from 27.5\% to 92.7\% by updating both modules by introducing head-to-head communication, which is newly introduced in Neuro-SERKET, between the LDA and ASR modules. 
The language model in the ASR is updated by referring to the probabilistic clustering result of the LDA module, and it is thought that the ASR outputs words that are relatively easy for the LDA module to cluster.
In addition, by sharing information between the GMM and LDA modules using the tail-to-tail module, which is also newly introduced in Neuro-SERKET, the performance of the GMM module was also significantly increased by approximately 25\%. The performance of the LDA module is also slightly increased.

\begin{figure}
\centering
 \includegraphics[width=0.8\linewidth]{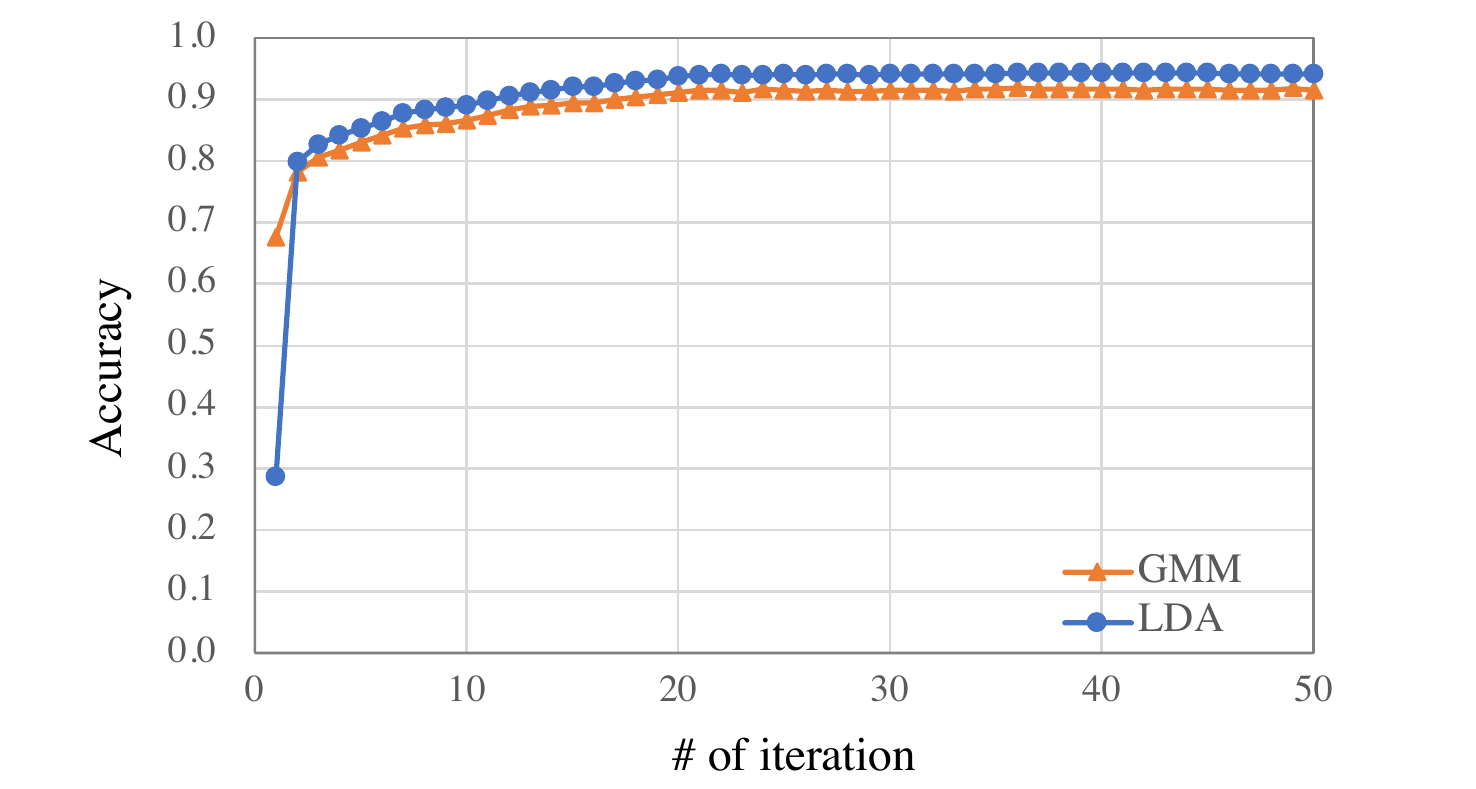}
\caption{Transition of classification accuracy. }
\label{fig:accuracy}       % Give a unique label
\end{figure}

Figure~\ref{fig:accuracy} shows the transition of the classification accuracy of VAE+GMM+LDA+ASR.
It shows that the performances of the LDA and GMM modules gradually increased.

\begin{figure}
\centering
 \includegraphics[width=\linewidth]{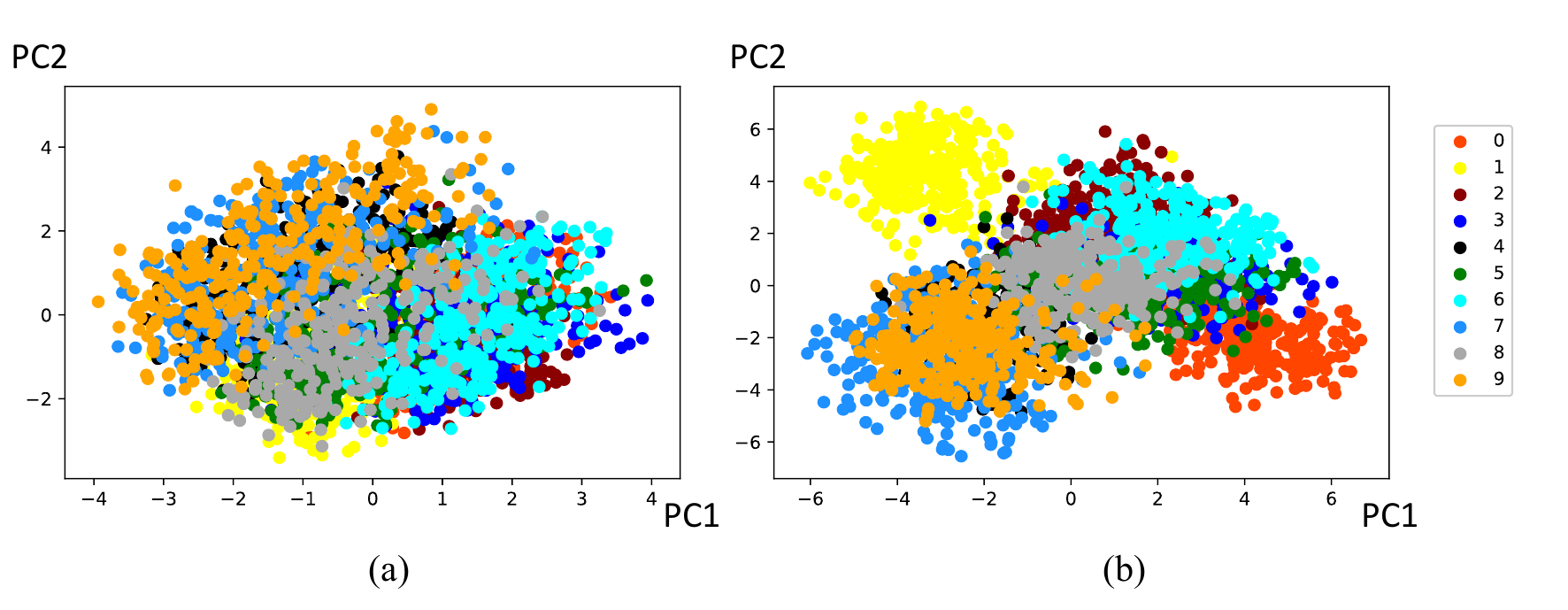}
\caption{Latent space of VAE learned using (a) VAE GMM LDA ASR  and (b) VAE+GMM+LDA+ASR. Proportion of variance for [PC1, PC2] are $[0.15, 0.13]$ and $[0.24, 0.20]$ for (a) and (b), respectively.}
\label{fig:latent_space} 
\end{figure}

Figure~\ref{fig:latent_space} shows the representations learned by the VAE. The ten-dimensional latent space of the VAE is compressed into a two-dimensional space using a principal component analysis (PCA) for visualization. Each color represents a digit.
Figure~\ref{fig:latent_space} (a) and (b) show the results of embedding without and with communication, i.e., without SERKET and with Neuro-SERKET, respectively.
This result shows that VAE+GMM+LDA+ASR formed an appropriate latent space for clustering using the GMM module.

Next, we observed clustered words. Each cluster involves numerous words having recognition errors. To check if each cluster corresponds to a digit, we picked up a stereotypical word, i.e., a syllable sequence, $\bar{\bs}_c$, by using the following equations.
   \begin{eqnarray}
   \bar{j}_c &=& {\rm argmin}_j \frac{ 1 }{I_c} \sum_i^{I_c} D( \bs_{cj}, \bs_{ci} ) \\
   \bar{\bs}_c &=& \bs_{c \bar{j} }
   \end{eqnarray}
where $I_c$ is the number of words classified into class $c$; $\bs_{ci}$ is the $i$-th word, i.e., the syllable sequence, classified into class $c$; and $D(\cdot, \cdot)$ represents the edit distance between the two-syllable sequence.
This procedure selects a word that is nearest to the center of the set of words in terms of the edit distance. Therefore, we can consider $\bar{\bs}_c$ as a stereotype of class $c$. 
The determined stereotypes of each class are shown in Table~\ref{tbl:average_words}.
Compared with Table~\ref{tbl:japanese_number}, we can see that unsupervised learning using VAE+GMM+LDA+ASR can acquire an appropriate syllable sequence for each number.

\begin{table}
\caption{Stereotypical word in each class. The italic characters denote errors. }
\label{tbl:average_words}
\centering
\begin{tabular}{lclcl} \hline
Number & Japanese pronunciation  \\ \hline
0 & ze ro \\ \hline
1  & i chi {\it i} \\ \hline
2 & ni {\it n i} \\ \hline
3 & sa n \\ \hline
4 & yo n \\ \hline
5 & go {\it o} \\ \hline
6 & ro ku \\ \hline
7 & na {\it n} na {\it a} \\ \hline 
8 & ha chi \\ \hline
9 & kyu u \\ \hline
\end{tabular}
\end{table}

\section{Conclusion}\label{sec:5}
To develop an integrative cognitive system using PGMs more efficiently, we require a useful framework that allow us to reuse elemental cognitive modules developed by other researchers and developers. This paper described Neuro-SERKET, which is a framework for developing a complex cognitive system by composing elemental PGMs. Neuro-SERKET is an extension of SERKET, which can compose elemental PGMs developed in a distributed manner. Although SERKET only supports a head-to-tail connection, Neuro-SERKET supports tail-to-tail and head-to-head connections. In addition, Neuro-SERKET supports neural network-based modules, e.g., deep generative models such as VAEs, which are not supported by conventional SERKET. As an example application of Neuro-SERKET, an integrative model called VAE+GMM+LDA+ASR was developed by composing VAE, GMM, LDA, and ASR. The performance of VAE+GMM+LDA+ASR and the validity of Neuro-SERKET are demonstrated through a multimodal categorization task using image data and the speech signal of numerical digits.     

In this paper, we showed only one example, i.e., VAE+GMM+LDA+ASR, and demonstrated the validity of Neuro-SERKET. Further application of Neuro-SERKET and the development of cognitive systems that enable a robot to form concepts, learn behaviors, and acquire language in a real-world environment is our future challenge. 
In particular, it has become clear that language learning in a real-world environment requires a wide range of cognitive capabilities~\cite{taniguchi2019langrobo}.
For this reason, at least two additional approaches should be applied for Neuro-SERKET. 

The first one is the development of a software environment, i.e., software libraries. Nakamura et al. has been developing SERKET\footnote{SERKET: \url{http://serket.naka-lab.org/}}.  As described in this paper, the Neuro-SERKET framework fully includes the conventional SERKET framework. Therefore, the SERKET software environment should be naturally extended to the Neuro-SERKET software environment. To involve DGMs into the SERKET framework, making use of a pre-existing software library for the DGMs may be a reasonable solution. In addition, Suzuki et al. have been developing Pixyz\footnote{Pixyz: \url{https://github.com/masa-su/pixyz}}, which is a framework for DGMs. We are now working on an efficient utilization of Pixyz for Neuro-SERKET. We consider it is important to combine unsupervised learning by probabilistic models and representation learning by NNs such as VAEs, i.e., DGMs. 
As shown in this paper, the latent space suitable for classification can be learned by the interaction between them. 
We have also proposed the method for motion segmentation where GP-HSMM, which is a probabilistic model, and VAE, which can extract features from motions, are connected and we showed that low dimensional features suitable for segmentation can be learned by the interaction between them. 
We consider such a composed model of unsupervised learning and representation learning has the potential to solve the various problem and it is possible to construct these models easily by Neuro-SERKET  

The second is an exploration of the applicability of Neuro-SERKET. In the current version, the Neuro-SERKET framework heavily relies on a PoE approximation. The limitation of a PoE approximation should be investigated both theoretically and empirically. A series of studies forming the background of Neuro-SERKET are developing cognitive systems that can perform life-long learning in a real-world environment. Such a learning process involves behavioral learning and language acquisition. For this purpose, the system will receive unstructured sensorimotor data. Theoretical and empirical validations should be applied for further applications. So far, many researchers, including the authors, have proposed a lot of cognitive models for robots: object concept formation based on its appearance, usage and functions \cite{tool_model}, formation of integrated concept of objects and motions \cite{mmlda}, grammar learning \cite{grammar}, language understanding \cite{lang_und}, spatial concept formation and lexical acquisition\cite{taniguchi2016spatial,taniguchi2017online,Ishibushi2015}, simultaneous phoneme and word discovery~\cite{taniguchi2016nonparametric,taniguchi2016double,nakashima2019unsupervised}  and cross-situational learning~\cite{taniguchi2017cross,Amir2017}. 
These models are regarded as an integrative model that are constructed by combining small-scale models. Therefore, they can be also re-implemented by using Neuro-SERKET more efficiently.

The computational efficiency needs to be improved as well. 
The most of modules are implemented using pure python without parallel computation in current Neuro-SERKET except for VAE, which is implemented using TensorFlow. Therefore, parameter estimation is not so fast. The parameter estimation in independent modules can be parallelized, and it might be faster by implementing using C language and TensorFlow. 
We plan to improve these drawbacks in the future. Regarding an optimization policy, we manually set the order of modules to be updated in the experiment. 
However, we also found that the performance of the whole model changed depending on the order of modules to be updated. Therefore, we will study and create the guideline about the order of the models to be updated for the practical use of SERKET.

Neuro-SERKET allows us to focus on the integration and exploration of complex cognitive systems. Recently, multimodal learning with DGMs has been gaining attention. However, as the cerebral cortex in our human brain demonstrates, the human cognitive system is based on mutually connected cortical areas, which are considered to have respective functions and modality-dependent information processing. Doya hypothesized that the cerebral cortex is trained simply through unsupervised learning~\cite{doya1999}. In general, unsupervised learning is modeled by PGMs. Neuro-SERKET enables us to explore a constructive model of the cerebral cortex using PGMs. Such exploration and the development of a brain-inspired whole-brain cognitive architecture are also future challenges. We believe that Neuro-SERKET will be a key framework for the future constructive studies on general intelligence and symbol emergence in natural and artificial cognitive systems~\cite{Taniguchi2016symbol,Taniguchi2018symbol}.

\begin{acknowledgements}
This research was supported by MEXT/JSPS KAKENHI, Grant Nos. 16H06569 in \#4805 (Correspondence and Fusion of Artificial Intelligence and Brain Science) and 18H03308,
 and JST CREST (JPMJCR15E3).
\end{acknowledgements}

\bibliographystyle{spphys}       % APS-like style for physics
\bibliography{neuro-serket}

% \begin{thebibliography}{}
% %
% % and use \bibitem to create references. Consult the Instructions
% % for authors for reference list style.
% % Format for Journal Reference
% Author, Article title, Journal, Volume, page numbers (year)
% Author, Book title, page numbers. Publisher, place (year)
% \end{thebibliography}

\end{document}